\documentclass{article} %
\usepackage[nonatbib, preprint]{neurips_2021}
\usepackage{times}

\usepackage[numbers]{natbib}

\usepackage{amsmath,amsfonts,bm}

\def\eqref#1{equation~\ref{#1}}

\def\1{\bm{1}}

\DeclareMathAlphabet{\mathsfit}{\encodingdefault}{\sfdefault}{m}{sl}
\SetMathAlphabet{\mathsfit}{bold}{\encodingdefault}{\sfdefault}{bx}{n}

\newcommand{\E}{\mathbb{E}}

\usepackage{mdpn}
\usepackage{todonotes}

\usepackage[utf8]{inputenc} %
\usepackage[T1]{fontenc}    %
\usepackage{hyperref}       %
\usepackage{url}            %
\usepackage{booktabs}       %
\usepackage{amsfonts}       %
\usepackage{nicefrac}       %
\usepackage{microtype}      %
\usepackage{bm,bbm,amsmath,amsfonts,amssymb}
\usepackage{xspace}
\usepackage{xcolor}
\usepackage{eucal}
\usepackage{amsmath}
\usepackage{mathtools}  %
\usepackage{appendix}

\usepackage{enumitem}

\usepackage{algorithm}
\usepackage[ruled,vlined,algo2e,linesnumbered]{algorithm2e}

\usepackage{hyperref}

\usepackage[center]{caption}
\usepackage{subcaption}
\usepackage{wrapfig}

\usepackage{amsthm}	%

\usepackage{theoremref} %
\usepackage{thmtools}
\usepackage{thm-restate} %
\usepackage{amsxtra} %
\usepackage{hyperref}
\usepackage{url}

\usepackage{wrapfig}

\usepackage{textcase}

\usepackage{breqn}

\newcommand{\wic}{\textsc{wic}}
\newcommand{\vic}{\textsc{vic}}
\newcommand{\rvic}{\textsc{rvic}}
\newcommand{\diayn}{\textsc{diayn}}

\title{Wasserstein Distance Maximizing Intrinsic Control}

\author{Ishan Durugkar \thanks{work done while at DeepMind}  \\
Department of Computer Science\\
The University of Texas at Austin\\
Austin, TX, USA 78703 \\
\texttt{ishand@cs.utexas.edu} \\
\And
Steven Hansen, Stephen Spencer, Volodymyr Mnih \\
DeepMind
}

\begin{document}

\maketitle

\begin{abstract}
This paper deals with the problem of learning a skill-conditioned policy that acts meaningfully in the absence of a reward signal.
Mutual information based objectives have shown some success in learning skills that reach a diverse set of states in this setting.
These objectives include a KL-divergence term, which is maximized by visiting distinct states even if those states are not far apart in the MDP.
This paper presents an approach that rewards the agent for learning skills that maximize the Wasserstein distance of their state visitation from the start state of the skill.
It shows that such an objective leads to a policy that covers more distance in the MDP than diversity based objectives, and validates the results on a variety of Atari environments.
\end{abstract}

\section{Introduction} \label{sec:intro}

This paper considers the unsupervised reinforcement learning problem of learning a set of skill-conditioned policies that act meaningfully in an environment in the absence of an extrinsic reward signal.
Some previous works \citep{gregor2016variational, eysenbach2018diversity, baumli2021relative} approached this problem by using a mutual information objective to maximize the empowerment of the skill-conditioned policies.
In essence, such a mutual information objective is maximized by learning goal-conditioned policy and a discriminator such that the discriminator can infer which skill was executed by considering the states visited by the policy conditioned on that skill.

This type of objective has been shown to learn diverse skills which can be useful for exploration and heirarchical reinforcement learning (HRL) \citep{eysenbach2018diversity,baumli2021relative}.
However, one potential issue with mutual information-based objectives is that they can learn skills that are discriminable but do not move far from the agent's starting state \cite{campos2020edl}. 

This paper instead presents an approach which considers the Wasserstein distance between the state visitation distribution of the agent's skill-conditioned policy and its start state distribution and trains the agent to maximize this distance, an approach we term Wasserstein distance maximizing Intrinsic Control (\wic).
\wic\ also encourages the learning of diverse skills by constructing the reward function to prefer each skill maximizing the Wasserstein distance in a unique direction.
We hypothesize that maximizing the Wasserstein distance will lead to policies that cover more distance in the underlying environment.
This hypothesis is validated on two grid world environments where the policy learned using \wic\ maximizes the number of states that it visits, whereas \vic\ and related techniques are content to reach states that are discernible from each other.

Finally, we end with some preliminary results on the Atari benchmark that suggest that \wic\ is a promising approach to unsupervised intrinsic control.

\section{Related Work} \label{sec:related}

Intrinsic motivations \citep{baldassarre_intrinsic_2013, oudeyer2009intrinsic, oudeyer2008can} are rewards presented by an agent to itself in addition to the external task-specific reward.
Intrinsic motivation has been proposed as a way to encourage RL agents to learn skills \citep{barto2004intrinsically, barto2005intrinsic, singh2005intrinsically, santucci2013best} that might be useful across a variety of tasks, or as a way to encourage exploration \citep{bellemare2016unifying, csimcsek2006intrinsic, baranes2009r, forestier2017intrinsically}.
The optimal reward framework \citep{singh_intrinsically_2010, sorg2010internal} and shaped rewards \citep{ng1999policy} (if generated by the agent itself) also consider intrinsic motivation as a way to assist an RL agent in learning the optimal policy for a given task.
Such an intrinsically motivated reward signal has previously been learned through various methods such as evolutionary techniques \citep{niekum2010evolved, schembri2007evolving}, meta-gradient approaches \citep{sorg2010reward, zheng_learning_2018,zheng_what_2020}, and others.
The Wasserstein distance, in particular, has been used to present a valid reward for speeding up learning of goal-conditioned policies \citep{durugkar2021adversarial}, imitation learning \citep{xiao_wasserstein_2019, dadashi_primal_2020, zhang2020WDAIL}, as well as program synthesis \citep{ganin2018synthesizing}.

Mutual information based objectives have been used to learn skill-conditioned policies that act meaningfully in the absence of an external reward \citep{gregor2016variational,eysenbach2018diversity,campos2020edl,baumli2021relative}.
This paper considers the same problem but uses the Wasserstein distance as the objective the agent seeks to maximize.

\section{Background and Setup}
\label{sec:background}

In this section we set up the problem and go over some of the concepts relevant to the setting.

\subsection{Problem Setting}
Our environment is a special case of Markov decision processes without a reward function denoted by the tuple $ \mathcal{M}: \langle\mathcal{S}, \mathcal{A}, \mathcal{P}, \mu \rangle$.
$\mathcal{S}$ is the state space, $\mathcal{A}$ is the action space, $\mathcal{P}: \mathcal{S} \times \mathcal{A} \longmapsto \bigtriangleup(\mathcal{S})$ is the conditional distribution denoting the transition dynamics when taking action $a \in \mathcal{A}$ from state $s \in \mathcal{S}$ ($\bigtriangleup$ denotes a distribution over the set given as the argument), and $\mu: \bigtriangleup(\mathcal{S})$ is the initial state distribution.

Agents interact with the environment with a skill conditioned policy $\pi_\theta: \mathcal{S} \times \Omega \longmapsto \bigtriangleup(\mathcal{A})$ where $\Omega$ is the space of skills, and $\theta$ denotes the parameters of the policy.
We assume that skills are sampled with some probability $P(\omega)$ which we assume to be fixed as a uniform distribution over a discrete set of skills in this paper.
In particular, we assume skills are sampled uniformly from set $\Omega$ and are then followed for a fixed number of $T$ time steps.
A skill-episode starts when a skill to be followed $\omega \in \Omega$ is sampled from the skill distribution, and continues for $T$ time steps.
The trajectory of states and actions that are obtained while the agent executing skill $\omega$ interacts with the environment is denoted by $s_0, a_0, s_1, \ldots, s_{T-1}, a_{T-1}, s_T$.
$s_T$ at the end of one skill episode acts as the first state $s_0$ for the next skill episode.
We will refer to $s_0$ and $s_T$ as the start state and end state of a skill episode respectively.

Finally we define the state visitation distribution of the policy $\pi_\theta$ conditioned on skill $\omega$ and starting at state $s_0$ as:
\begin{align} \label{eqn:sdist}
    \rho_\theta(s|s_0, \omega) = \frac{1}{T}\sum_{t=1}^T P(s_t=s|\pi_\theta, s_0, \omega)
\end{align}

\subsection{Intrinsic Control by maximizing Mutual Information}
 Variational Intrinsic Control (VIC) \citep{gregor2016variational} takes the above setting and sets up the problem of learning the skill-conditioned policy as one of maximizing the mutual information between the random variable $\omega$ denoting the skill and the states $S_T$ reached after executing the skill conditioned policy $\pi_\theta$ conditioned on $\omega$.
 Practically, this approach is implemented by learning a discriminator $D_\phi(\omega| s_T, s_0)$ with parameters $\phi$ that tries to predict the skill the agent policy was conditioned on given the start and end states of the skill-episode.
 
 The output of this discriminator is then used as the reward signal to train the skill-conditioned policy $\pi_\theta$.
 This approach encourages the learning of skills that can be distinguished by the end states of their trajectories.
 However, it does not encourage the learning of skills that travel as far as possible in the environment.
 
 Other works that utilize this mutual information objective use a similar setup, but encourage the discriminator to predict the skill based on the relative direction of the states reached compared to the start state (\rvic, \citet{baumli2021relative}) or try to discriminate the entire trajectory (\diayn, \citet{eysenbach2018diversity}).
 
\subsection{Wasserstein Distance and Optimal Transport}
The field of optimal transport \citep{peyre_computational_2020} considers the question of how to transport one distribution to another while minimizing the amount of effort expended.
The Wasserstein distance estimates the amount of work that needs to be done to convert one probability distribution to the other, as measured by the ground metric $d$.
More concretely, consider a metric space $(\mathcal{M}, d)$ where $\mathcal{M}$ is a set and $d$ is a metric on $\mathcal{M}$.
The Wasserstein-$p$ distance between two distributions $\mu$ and $\nu$ on $\mathcal{M}$ with finite moments can be defined as:
\begin{align}\label{eqn:wp}
    W_d^p(\mu, \nu) := \min_{\zeta \in Z} \mathbb{E}_{x,y \sim \zeta} \left[d(x, y)^p \right]^{1/p}
\end{align}
where $Z$ is the space of joint distributions $\zeta \in \bigtriangleup(\mathcal{M} \times \mathcal{M})$ whose marginals are $\mu$ and $\nu$ respectively.

While prior work on using the Wasserstein distance in reinforcement learning has used Euclidean distance in the state space as the ground metric~\cite{ganin2018synthesizing}, it may not be appropriate since it does not reflect the structure of the reinforcement learning problem. In MDPs, if we consider the metric space on the set of states $\mathcal{S}$, then the number of time-steps it would take to go from state $x$ to state $y$ under the current agent policy $\pi$ is a quasimetric (metric that might not be symmetric between $x$ and $y$) that could be considered more appropriate for measuring this work \citep{durugkar2021adversarial, jevtic2018combinatorial} since it reflects distance in the MDP instead of distance in observation space.

If we restrict our attention to the Wasserstein-1 metric, the Kantorovich-Rubinstein duality allows us to express the Wasserstein-1 distance (which we refer to simply as the Wasserstein distance hereafter) in the following manner:
\begin{align} \label{eqn:W_dual}
    W_d^1(\mu, \nu) = \sup_{Lip(f) \leq 1} \mathbb{E}_{y \sim \nu} \left[f(y)\right] - \mathbb{E}_{x \sim \mu} \left[ f(x) \right]
\end{align}

where the supremum is over all $1$-Lipschitz functions $f : \mathcal{M} \longmapsto \mathbb{R}$ in the metric space.
If this metric space is based on the time-step metric alluded to above, then the Lipschitz constraint can be enforced using the following equation for Lipschitz smoothness based on transitions experienced by the policy \citep{durugkar2021adversarial}:
\begin{align} \label{eqn:f_cons}
    Lip(f) = \max_{s \in \mathcal{S}} \mathbb{E}_{s' \sim \pi, \mathcal{P}} \left[ |f(s') - f(s)| \right]
\end{align}

The Wasserstein distance between two distributions can then be estimated by means of a function approximator such as a neural network \citep{arjovsky_wasserstein_2017, gulrajani_improved_2017, durugkar2021adversarial} by solving for equation \ref{eqn:W_dual} and ensuring smoothness according to Equation \ref{eqn:f_cons}.
In Section \ref{sec:WIC}, we lay out the exact objective to train such a function approximator.

\section{Wasserstein Distance Maximizing Intrinsic Control} \label{sec:WIC}

This section describes how the Wasserstein distance can be used to learn a skill conditioned policy, an approach we term Wasserstein distance maximizing Intrinsic Control (\wic).
At a high level, it proposes a method to learn a skill conditioned policy that attempts to get as far away from the skill's start state as measured through transitions in the MDP, and attempt to go in unique directions for each skill.
That is, \wic\ will train a policy to maximize $ \mathbb{E}_{\omega \sim P(\cdot | \Omega)} \left[W_d^1(\delta(s_0), \rho_\theta(s|s_0, \omega))\right]$, and penalize this policy for maximizing $\mathbb{E}_{\omega' \neq \omega } \left[W_d^1(\delta(s_0), \rho_\theta(s|s_0, \omega'))\right]$ for any other skill $\omega' \neq \omega$.

For a particular skill $\omega \in \Omega$ that starts executing at a state $s_0$, the above Wasserstein distances are estimated between a Dirac distribution at the skill's start state $\delta(s_0)$ and the skill's state visitation distribution $\rho_\theta(s|s_0, \omega)$ (Equation \ref{eqn:sdist}) with a potential function $f_\phi : \mathcal{S} \times \mathcal{S} \times \Omega \longmapsto \mathbb{R}$ with parameters $\phi$.
The potential function $f_\phi$ is trained by minimizing the following objective:
\begin{align} \label{eqn:l_obj}
    L_f(s_0, \rho_\theta, \omega) = f_\phi(s_0, s_0, \omega) - \mathbb{E}_{s \sim \rho_\theta(\cdot |s_0, \omega)} \left[f_\phi(s, s_0, \omega)\right]
\end{align}
while also enforcing that the potential function $f_\phi$ is $1$-Lipschitz with the following objective:
\begin{align} \label{eqn:l_cons}
    L_c(s_0, \rho_\theta, \omega) = \E_{s, s' \sim \rho_\theta(\cdot |s_0, \omega)}\text{maximum}(\|f_\phi(s', s_0, \omega) - f_\phi(s, s_0, \omega) \|^2 - 1, 0)
\end{align}
where $s$ is a state drawn according to the state visitation distribution $\rho_\theta$ and $s'$ is a sample of the next state the agent would visit if following policy $\pi_\theta$ conditioned on skill $\omega$ in that state.
Maximizing the Wasserstein distance thus estimated will lead to a skill that attempts to get as far away from the start state as possible.

In order to maximize this distance, the agent is trained with rewards that encourage it to move its state visitation distribution to regions of higher potential, and thus increase the Wasserstein distance of the state visitation distribution.
Since the potential function is state-based, it is enough for the reward to be a difference in potentials \citep{ng1999policy}.
Further, \wic\ also includes a term to encourage diverse skills, meaning ones that move in unique directions in the state space.
This diversity is encouraged by including a penalty term for overlapping with the positive potential gradient of any other skill.
Consequently, the reward we present to the agent is as follows:
\begin{align}
    \label{eqn:rew}
    r(s_t, a_t, s_{t+1}, s_0, \omega) := &\left[f(s_{t+1}, s_0, \omega) - f(s_{t}, s_0, \omega)\right] - \nonumber\\ &\eta \max_{\omega' \neq \omega} \left[f(s_{t+1}, s_0, \omega') - f(s_{t}, s_0, \omega')\right]
\end{align}
where $\eta \in [0, 1]$ specifies how much of a penalty skills get for encouraging a direction that overlaps with other skills.

\section{Experiments}

We compare \wic\ with \vic\ in domains with increasing order of complexity in order to probe the difference between the skills learned by maximizing the Wasserstein distance to its start state as opposed to maximizing a mutual information objective with respect to a fixed skill sampling distribution.

\begin{figure} [t]
\centering
\begin{subfigure}[t]{0.24\textwidth}
\centering
\includegraphics[width=\linewidth]{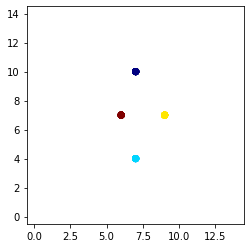}
\caption{End points of skills learned by \vic}
\label{fig:vic_end}
\end{subfigure}
\hfill
\begin{subfigure}[t]{0.24\textwidth}
\centering
\includegraphics[width=\linewidth]{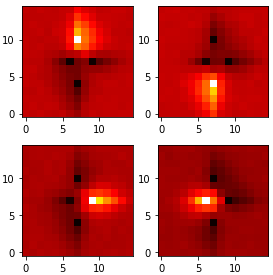}
\caption{Rewards learned by \vic}
\label{fig:vic_rew}
\end{subfigure}
\hfill
\begin{subfigure}[t]{0.24\textwidth}
\centering
\includegraphics[width=\linewidth]{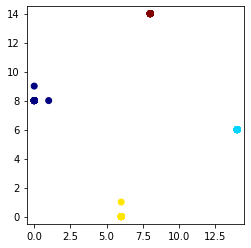}
\caption{End points of skills learned by \wic}
\label{fig:wic_end}
\end{subfigure}
\hfill
\begin{subfigure}[t]{0.24\textwidth}
\centering
\includegraphics[width=\linewidth]{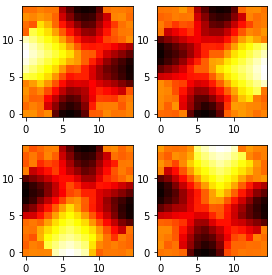}
\caption{Rewards learned by \wic}
\label{fig:wic_rew}
\end{subfigure}
\caption{The endpoints reached after executing all the skills multiple times starting in the middle of the room, and reward functions learned for \vic\ and \wic\ respectively in a $15 \times 15$ grid world where the features are a one-hot encoding. The agent starts executing each skill from the center of the grid world.}
\label{fig:tab}
\end{figure}

\subsection{Tabular}
First, we evaluate \textsc{wic} and \textsc{vic} on a tabular grid world domain.
The grid is $15 \times 15$ giving us 225 distinct states with $5$ possible actions (up, down, left, right, and no-op), and the agent starts off in the center of the grid.
Since this domain is tabular, the agent's state is communicated as a one-hot vector.

There are no obstacles, and $|\Omega| = 4$ skills which are sampled uniformly.
Once the skill $\omega$ is sampled, the agent executes its skill-conditioned policy for $T=10$ time steps.
After executing this policy for $T$ time steps, the state is reset back to the center, a new skill is sampled, and the policy executes again for $T$ time steps.
We compare two methods to learn and present the intrinsic reward, \vic\ and \wic.
Both the agent policy and the discriminator (\vic) or potential function (\wic) are linear functions of the features.
For \wic, a penalty $\eta=0.9$ is used to avoid learning skills that overlap in their state visitation.

The agent policy is trained using REINFORCE \citep{williams1991probability} with a state-conditioned baseline.
The policy is additionally regularized with an entropy loss weighted by $0.01$ to prevent premature convergence.
Both the discriminator and the policy are trained using stochastic gradient descent \citep{robbins1951stochastic, kiefer1952stochastic} with a learning rate of $0.003$.

Figure \ref{fig:tab} shows the states reached after executing each skill multiple times from the same start state at the middle of the room (Figure \ref{fig:vic_end}) and the reward function used to train the policy (Figure \ref{fig:vic_rew}) for \vic.
Figures \ref{fig:wic_end} and \ref{fig:wic_rew} show the same respectively for \wic.
As hypothesized, \vic\ is satisfied with reaching states that are distinct enough for the discriminator to tell apart, and does not necessarily attempt to learn a policy that travels far in the environment.
\wic\ on the other hand, learns a reward function and a policy that attempt to travel as far away from the start state as possible.

\begin{figure} [t]
\centering
\begin{subfigure}[t]{0.24\textwidth}
\centering
\includegraphics[width=\linewidth]{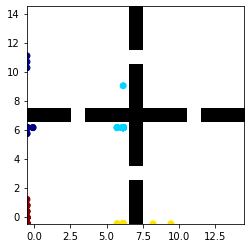}
\caption{End points of skills learned by \vic}
\label{fig:vic_end_4rooms}
\end{subfigure}
\hfill
\begin{subfigure}[t]{0.24\textwidth}
\centering
\includegraphics[width=\linewidth]{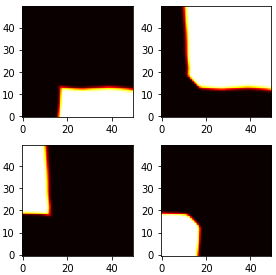}
\caption{Rewards learned by \vic}
\label{fig:vic_rew_4rooms}
\end{subfigure}
\hfill
\begin{subfigure}[t]{0.24\textwidth}
\centering
\includegraphics[width=\linewidth]{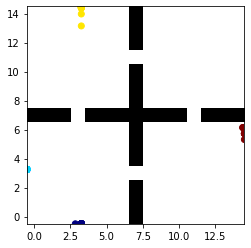}
\caption{End points of skills learned by \wic}
\label{fig:wic_end_4rooms}
\end{subfigure}
\hfill
\begin{subfigure}[t]{0.24\textwidth}
\centering
\includegraphics[width=\linewidth]{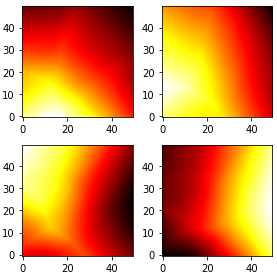}
\caption{Rewards learned by \wic}
\label{fig:wic_rew_4rooms}
\end{subfigure}
\caption{Visualizing the endpoints after executing all the skills multiple times starting in the middle of the bottom-left room, and the reward functions used to learn them by \vic\ and \wic\ respectively in the Four Rooms environment. The features of the state here are the $(x, y)$ coordinates of the point at which the agent is. The agent starts executing each skill from the center of the bottom left room.}
\label{fig:4rooms}
\end{figure}

\subsection{Four Rooms}
Next, we evaluate how \wic\ and \vic\ behave differently when the features are not one-hot vectors.
We use a four room domain with the location of the agent communicated as its $(x, y)$ coordinate, and each feature scaled to $[-1, 1]$.
The agent starts in the center of the bottom left room and each skill is allowed $T=40$ time steps to execute.
This duration is enough for an agent to make it to the room diagonally opposite if the skill-conditioned policy is deterministic.

The number of skills $|\Omega|=4$ and they are sampled uniformly.
After the agent finishes executing one skill it samples a new skill and begins executing it from the state that was reached.
The agent is reset to the middle of the bottom left room after sampling and executing skills $17$ times.

Both the policy and the potential function (\wic) or discriminator (\vic) are instantiated as multi-layer perceptrons with $2$ hidden layers of $128$ units each, and ReLU \citep{nair2010rectified} activation functions internally.
The other training details remain the same as in the tabular case, except for the use of the Adam optimizer \citep{Kingma2015Adam} with its default learning rate of $0.001$.

The end states reached after executing the skill-conditioned policy for each skill multiple times from the middle of the bottom-left room are shown in Figure \ref{fig:4rooms}.
Here again, we see that the discriminative objective of \vic\ is satisfied with learning skills that go to the corners of the bottom left room, whereas \wic\ learns a policy that travels deep into the adjoining two rooms.
The reward functions visualized in both these domains makes it clear that the Wasserstein distance maximizing approach pushes the agent to go as far from the start state as it can.

\begin{figure} [bh!]
\centering
\begin{subfigure}[t]{\linewidth}
\centering
\includegraphics[width=\linewidth]{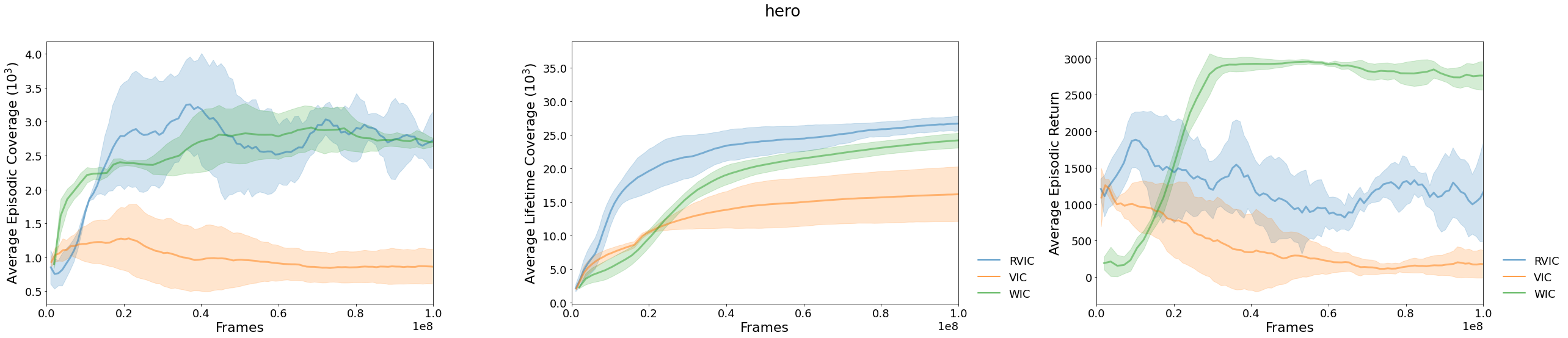}
\label{fig:hero}
\end{subfigure}
\\
\begin{subfigure}[t]{\linewidth}
\centering
\includegraphics[width=\linewidth]{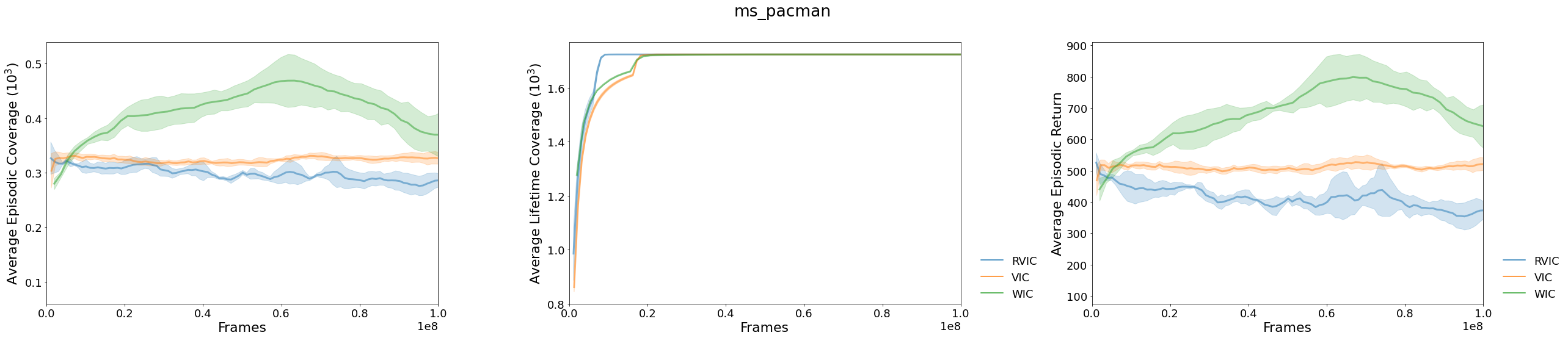}
\label{fig:pacman}
\end{subfigure}
\\
\begin{subfigure}[t]{\linewidth}
\centering
\includegraphics[width=\linewidth]{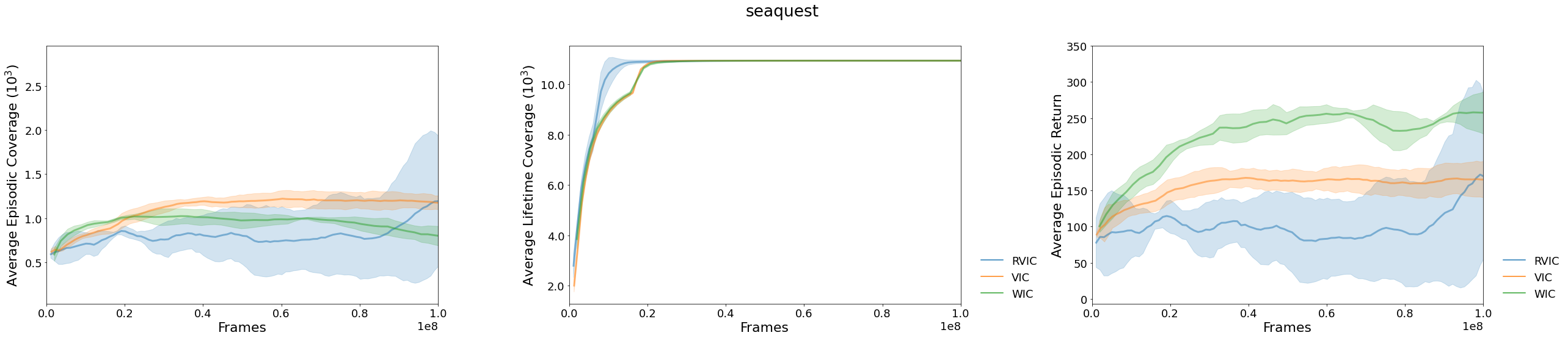}
\label{fig:seaquest}
\end{subfigure}
\\
\begin{subfigure}[t]{\linewidth}
\centering
\includegraphics[width=\linewidth]{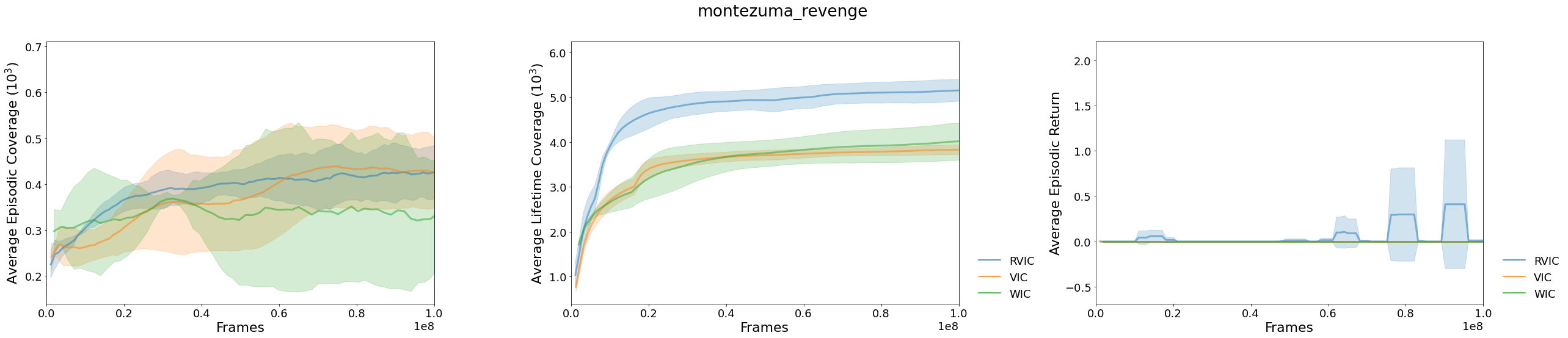}
\label{fig:montezuma}
\end{subfigure}
\\
\begin{subfigure}[t]{\linewidth}
\centering
\includegraphics[width=\linewidth]{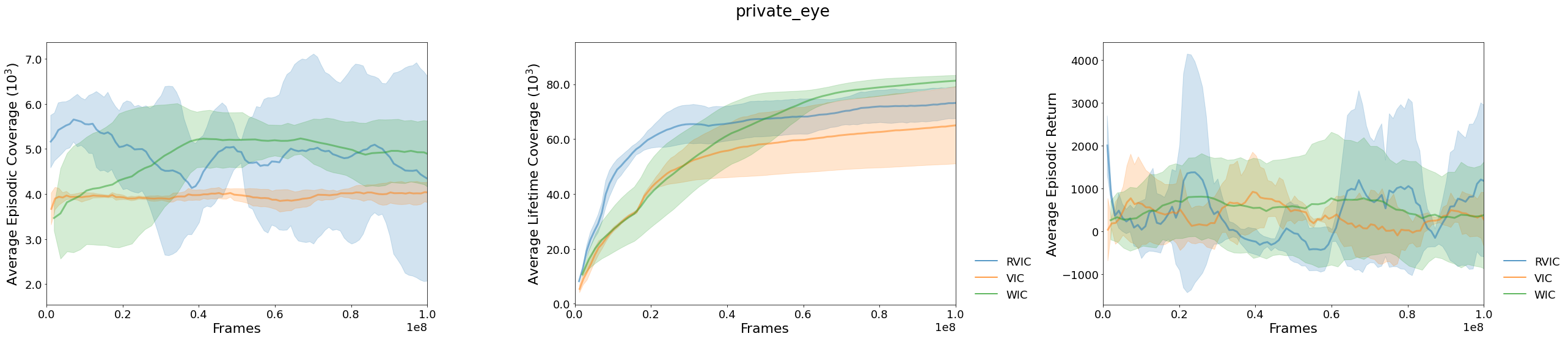}
\label{fig:private_eye}
\end{subfigure}
\\
\begin{subfigure}[t]{\linewidth}
\centering
\includegraphics[width=\linewidth]{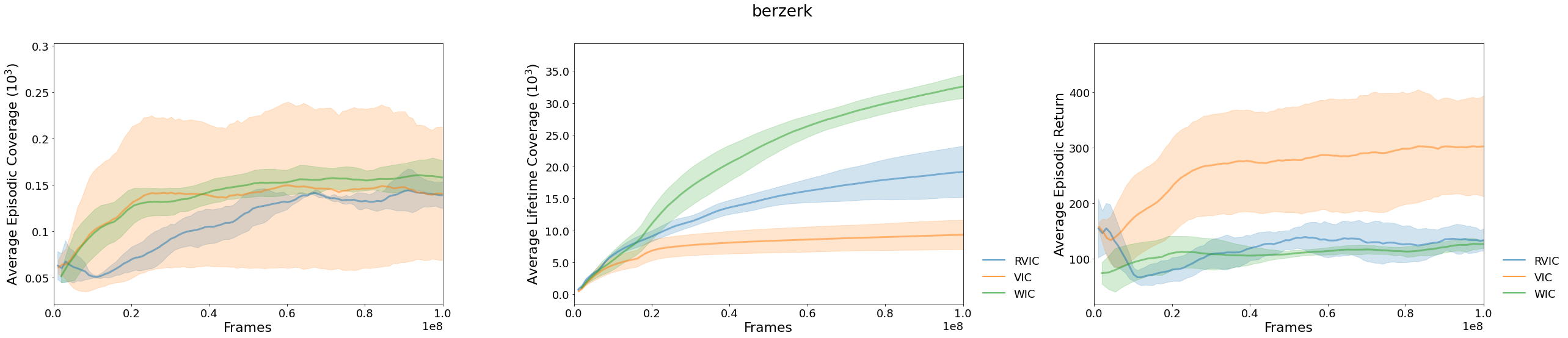}
\label{fig:berzerk}
\end{subfigure}
\caption{Comparing how well \wic\ does compared to mutual information based methods in Atari domains}
\label{fig:atari}
\end{figure}

\subsection{Atari}

So far, we have validated that \wic\ encourages the learning of skill-conditioned policies that try to go as far away from the state that the skill was invoked.
We now apply \wic\ to the Atari domain \citep{bellemare2013arcade, machado2018revisiting} and evaluate how well this approach scales to image based inputs and deeper function approximators.
\wic\ is compared to \vic\ and \rvic, and the metrics we use for comparison are average episodic coverage, average lifetime coverage, and average episodic return.

The potential function (\wic) or discriminator (\rvic\ and \vic) are trained just as before, but the agent's policy is now instantiated as a Q-function trained using Q($\lambda$) with $\lambda=0.9$ and discount factor $\gamma=0.98$.
A replay buffer of size $5 \times 10^5$ is used to store the data, and each minibatch samples $64$ trajectories of length $T=40$ from the buffer to train from.
The Q-function and the potential function share a common torso in this setting, and the architecture of this torso is equivalent to the one suggested in IMPALA \citep{espeholt2018impala}.

As can be seen from Figure \ref{fig:atari}, the skill-conditioned policy learned through \wic\ leads to episodic returns better than \vic\ or \rvic\ on three of the six games we test on: Hero, Montezuma's Revenge, and Ms. Pacman, and roughly equivalent returns on the other three.
These improved returns are indicative of more directed policies, even in games where the episodic coverage is similar to the mutual information based approaches (Hero and Seaquest).
In the game Berzerk, we additionally see that even though the episodic returns do not outperform \vic, in terms of lifetime coverage \wic\ widens the gap over time.

\section{Conclusion}
This paper considers the question of unsupervised learning of skills in an environment, and hypothesizes that maximizing the Wasserstein distance from the start state distribution of a skill could lead to skill-conditioned policies that cover more distance in the underlying MDP than mutual information based approaches like \vic\ and \rvic.
This approach is crystallized as Wasserstein distance maximizing intrinsinc control (\wic), and the above hypothesis is validated on a tabular grid world as well as a continuous four rooms domain.
Finally, we have validated that the approach scales up to visual inputs and complex environments by evaluating it on the Atari domain.

\bibliographystyle{plainnat}
\bibliography{main}

\end{document}